# Residual Attention Single-Head Vision Transformer Network for Rolling Bearing Fault Diagnosis in Noisy Environments

Songjiang Lai

The Hong Kong Polytechnic University, Kowloon, Hong Kong, song-jiang.lai@connect.polyu.hk

Centre for Advances in Reliability and Safety, New Territories, Hong Kong

Tsun-Hin Cheung

The Hong Kong Polytechnic University, Kowloon, Hong Kong, tsun-hin.cheung@connect.polyu.hk

Centre for Advances in Reliability and Safety, New Territories, Hong Kong

Jiayi Zhao

Centre for Advances in Reliability and Safety, New Territories, Hong Kong, jiayi.zhao@cairs.hk

Kaiwen Xue

The Hong Kong Polytechnic University, Kowloon, Hong Kong, 22039555r@connect.polyu.hk

Centre for Advances in Reliability and Safety, New Territories, Hong Kong

Ka-Chun Fung

The Hong Kong Polytechnic University, Kowloon, Hong Kong, 21119532r@connect.polyu.hk

Centre for Advances in Reliability and Safety, New Territories, Hong Kong

Kin-Man Lam

The Hong Kong Polytechnic University, Kowloon, Hong Kong, kin.man.lam@polyu.edu.hk

Centre for Advances in Reliability and Safety, New Territories, Hong Kong

Rolling bearings are critical components in modern industrial machinery, significantly impacting the performance, longevity, and safety of equipment. Due to harsh operating conditions, such as high speeds and temperatures, rolling bearings are prone to malfunctions, leading to equipment downtime, economic losses, and safety risks. In this paper, the Residual Attention Single-Head Vision Transformer Network (RA-SHViT-Net) is proposed for fault diagnosis in rolling bearings. The vibration signal collected from rolling bearings is first transformed from the time domain to the frequency domain using Fast Fourier Transform (FFT). The RA-SHViT-Net model then leverages the Single-Head Vision Transformer (SHViT), which is adept at capturing local and global features from time-series signals. SHViT also offers a state-of-the-art balance between computational complexity and prediction accuracy and it has been demonstrated to achieve promising results in the field of computer vision. To enhance feature extraction, we introduce an Adaptive Hybrid Attention Block (AHAB) that combines channel and spatial attention mechanisms. The core building block of the RA-SHViT-Net is the Residual Attention Single-Head Vision Transformer Block, which consists of a Depthwise Convolution (DWConv) layer, a Single-Head Self-Attention (SHSA) layer, a Residual Feed-Forward-Network (Res-FFN) and an Adaptive Hybrid Attention Block (AHAB). This architecture is designed to comprehensively extract vibration signal features by considering the interdependencies among feature channels and spatial information based on the excellent feature extraction capabilities of SHViT. Additionally, each Single-Head Vision Transformer Block incorporates a Residual Feed-Forward-Network (Res-FNN) module, which uses residual connections to mitigate the vanishing gradient problem, enabling stable and efficient training of deep models. This design enhances the model's ability to learn complex representations and improves its generalization capabilities. The proposed RA-SHViT-Net was evaluated using the Case Western Reserve University (CWRU) dataset and the Paderborn University dataset. The results demonstrate that the RA-SHViT-Net outperforms state-of-the-art methods in terms of accuracy and robustness, particularly in scenarios involving complex and noisy environments. In addition, we designed multiple ablation studies to investigate the impact of different modules on the network's prediction performance. Overall, the RA-SHViT-Net provides a powerful tool for the early detection and classification of bearing faults, contributing to more reliable and efficient maintenance strategies in industrial applications.

## 1 INTRODUCTION

In modern industrial machinery, rolling bearings are vital components playing a critical role in the operation of mechanical rotating systems, which are prevalent in various industries such as petrochemical, power generation, and transportation [1]. Their condition significantly affects the performance, longevity, and safety of the equipment. Operating under harsh conditions, such as high speeds and temperatures, rolling bearings are prone to malfunctions that can lead to equipment downtime, economic losses, and safety risks [2,3]. Precision spindle bearings, particularly in machine tools, are essential for ensuring machining quality and production efficiency, further highlighting the importance of timely fault detection and health monitoring to maintain system stability and prevent unexpected failures.

Methodologies for fault diagnosis of rolling bearings are generally categorized into model-based and data-driven approaches. Many early studies focused on model-based techniques. For instance, Cococcioni et al. [4] evaluated the condition of rolling bearings by manually selecting features and designing classifiers. Miao et al. [5] introduced Feature Mode Decomposition (FMD), which employs an adaptive finite impulse response (FIR) filter to iteratively decompose signals and extract hidden fault features. Other classical algorithms, such as Bayesian networks [6], k-Nearest Neighbor (k-NN) [7], and Self-Organizing Map (SOM) networks [8], have also been widely applied in bearing fault diagnosis. These traditional methods often require specific prior knowledge and assumptions for model construction and feature processing.

Deep learning has significantly advanced fault diagnosis by enhancing feature extraction and adaptability in recognition algorithms. Liang and Zhou [9] developed a hierarchical model with transfer learning, achieving 94.59% accuracy for rolling bearings in complex systems. Lu et al. [10] introduced a deep neural network incorporating maximum mean discrepancy for cross-domain bearing fault identification. Tang et al. [11] proposed a Kalman filter-aided deep belief network for multi-sensor data, achieving over 98% accuracy. In another approach, wavelet transform was used to convert vibration signals into 2D spectrograms, which were then analyzed using CNNs for planetary gearbox bearing fault diagnosis [12]. Liu and Zhou [13] introduced an autoencoder-based recurrent neural network (RNN) for bearing fault diagnosis. An et al. [14] proposed a Long Short-Term Memory (LSTM) framework, which addresses vanishing gradient issues, although training time remains a challenge for long sequences. Researchers Singh et al. [15] proposed a method that integrates Graph Attention Network (GAT) and Long Short-Term Memory (LSTM) to capture spatial and temporal dependencies in sensor data for accurate bearing fault detection. A novel method that converts vibration signals into graph data for processing by graph neural networks, incorporating feature fusion and an ensemble learning strategy for accurate fault diagnosis is proposed by Wang et al. [16].

In 2017, Vaswani et al. [17] introduced the Transformer, a model that uses self-attention mechanism to quickly capture global features from time-series signals. This innovation revolutionized natural language processing by addressing the challenge of extracting long-range dependencies while maintaining the sequential order of data. By discarding recursive and convolutional structures, the Transformer allows for parallel sequence computation, significantly enhancing training efficiency. After Transformer achieved success in the field of NLP, researchers applied it and its variants to other fields including fault diagnosis of rolling bearings and achieved quite good results. Up to now, the series of models with Transformer as the backbone have gradually replaced the traditional CNN-based models and become the mainstream in the field of bearing fault diagnosis. Li et al. [18] proposed GLP-Transformer, a lightweight method combining convolutional and transformer techniques for efficient and accurate bearing fault diagnosis. Hou et al. [19] proposed Diagnosisformer, an efficient rolling bearing fault diagnosis method that enhances feature extraction and parallel computing abilities through an improved Transformer-based multi-feature parallel fusion model. AFMD-Transformer framework that combines adaptive feature mode decomposition optimized by an improved coati optimization algorithm and a Transformer neural network to achieve high accuracy and robustness in rolling bearing fault diagnosis is also proposed by Xiao et al. [20]. However, Transformer-based models excel in performance but face challenges such as lack of inductive bias, quadratic computational complexity, and high resource demands. Conventional multi-head self-attention (MHSA) Transformers exhibit substantial computational redundancy. General efficient vision Transformers typically employ a 4×4 patch embedding and a four-stage structure. This design introduces significant spatial redundancy in the early stages, increasing memory access costs. Furthermore, the use of MHSA incurs high memory access costs due to operations such as Layer Normalization and data movement (e.g., reshaping), which consume a large portion of runtime. These operations are primarily memory-bound, leading to higher memory access overhead. Additionally, such models often lack effective local representations. Standard Vision Transformer (ViT) models commonly tokenize inputs using a 16×16 stride-16 convolution, which limits local representation capabilities. To mitigate these issues, Yun et al. [21] proposed the Single-Head Vision Transformer with Memory Efficient Macro Design (SHViT), a novel Transformer architecture designed by using larger-stride patch embeddings, substituting early-stage attention layers with convolutions, and introducing a single-head attention module to reduce redundancy and improve performance, achieving state-of-the-art speed-accuracy trade-offs on diverse devices. Compared to previous Transformers, SHViT significantly reduces spatial redundancy. By employing a larger-stride patchify stem, SHViT decreases spatial redundancy, thereby reducing memory access costs and utilizing more compact token representations in the early stages, which enhances performance. Additionally, SHViT introduces the Single-Head Self-Attention (SHSA) module. This module combines global and local information in parallel, eliminating the computational redundancy associated with the multi-head mechanism while improving accuracy. The SHSA layer applies self-attention only to a subset of input channels, leaving the rest unchanged, thus reducing memory access costs. Moreover, an efficient macro design is another key feature of SHViT. It adopts a compact token representation that increases semantic density, allowing for more blocks to be stacked within the same computational budget, further enhancing performance. SHViT has demonstrated superior performance in various computer vision tasks. Therefore, inspired by the success of the SHViT model in computer vision research area, we propose a lightweight model select SHViT as the backbone for bearing fault diagnosis, aiming to reduce the computational complexity and simultaneously improve the prediction accuracy of the model

To effectively capture the interdependencies among feature channels and spatial information, we introduce the Adaptive Hybrid Attention Block (AHAB) as a fundamental component of the SHViT, forming the Residual Attention Single-Head Vision Transformer Block (RA-SHViT). By combining channel and spatial attention, the model can create a more comprehensive and precise representation of the signal. This dual attention mechanism ensures that the model not only focuses on the most relevant features but also considers their spatial distribution, leading to better discrimination between different faulty conditions. Besides, the combined attention mechanism makes the model more robust to variations in the input data. It can handle different types of faults and operating conditions more effectively, reducing the likelihood of false positives and false negatives. This combination of RA-SHViT Blocks is classified into 3 stages. In stage 1, RA-SHViT Blocks consist of Depthwise Convolution (DWConv) layer, Residual Feed-Forward-Network (Res-FFN) and Adaptive Hybrid Attention Block (AHAB) and the number of RA-SHViT Blocks is set $L_1$. The structure of RA-SHViT Block in stage 2-3 has an additional Single-Head Self-Attention (SHSA) layer compared with stage 1 and the number of RA-SHViT Blocks in stage 2-3 is set $L_2$ and $L_3$ respectively. For the successive RA-SHViT Block, an AHAB is employed to generate specific attention for each channel and spatial feature. The AHAB integrates a new attention mechanism that combines channel and spatial attention, enabling the capture of fine-grained spatial details and the effective highlighting of informative channels within feature maps. Additionally, we incorporate a novel Residual Feed-Forward-Network (Res-FFN) module into the RA-SHViT Block architecture by replacing the original Feed-Forward-Network (FFN). The (Res-FFN) module uses residual connections to address the vanishing gradient problem, thereby stabilizing and enhancing the efficiency of training deep models. The proposed model, named Residual Attention Single-Head Vision Transformer Network (RA-SHViT-Net), is validated and tested on the Case Western Reserve University (CWRU) dataset and Paderborn bearing dataset, with experimental results demonstrating its high efficiency and superior performance.

## 2 METHODOLOGY

### 2.1 Adaptive Hybrid Attention Block

The attention mechanism can significantly enhance the accuracy of detecting surface damage in rolling bearings, particularly for a series of defects that include microscopic scratches, cracks, and peeling. Given the unique operating mode of rolling bearings, the metal surfaces are subjected to repetitive contact stress, resulting in material fatigue or wear and damage aggravated by inadequate lubrication and pollutant invasion. Unlike typical surface defects, the average length of small-scale damage on the surface of rolling bearings is less than 500um, and ordinary transformer-based models cannot properly capture these defects. However, the model's performance may be improved using the Adaptive Hybrid Attention Block (AHAB) described as proposed in this study.

The AHAB framework is designed based on the Channal Attention Module (CBAM) proposed by Sanghyun et al. [22], shown in Figure 1, where the feature with vector shape $F \in \mathbb{R}^{C \times H \times W}$ is used as input. In the channel attention branch, the average pooling associated with max pooling layer are organized as parallel, and the feature map will be processed by CNN followed by activation function before concatenation. Spatial attention branch uses input vector directly as the input feature and two convolutional layers are employed to fuse and emphasize spatial information. The output features of the two sub-branches are normalized using Min-max normalization function before conjunction, and adaptable parameters are provided to make the stream more flexible. Therefore, the proposed AHAB framework allows spatial attention module to fetch features from the input feature unmediated, which will enhance the model performance in data collected with disturbance, especially in complex and noisy environments.

In bearing fault diagnosis, since certain features in the signal may be more indicative of faults than others, channel attention helps to identify and emphasize these relevant features. By focusing on the most meaningful channels, the model can better distinguish between normal operation and faulty conditions. Channel attention can also help suppress these noisy channels, thereby improving the signal-to-noise ratio and making the fault detection more reliable. Besides, the model with the channel attention can reduce the computational load and improve efficiency, which is crucial for real-time monitoring systems by concentrating on the most important features. The Channel Attention module is based on a convolutional neural network that builds multi-channel feature maps using a succession of convolutional layers with activation function to represent various feature representations of the input vectors. In rolling bearing object fault diagnosis project, these features represent important information related to the defect itself or non-important information associated with irrelevant features such as noise. Using CAM to obtain input vectors will generate multiple feature maps, while obtaining the weight vector for each feature. For noise and disturbance, the relevant weight vectors will be weakened in multiple model iterations to avoid interference from high noise information of rolling bearings in the model.

Furthermore, the model proposed in this paper directly links the Spatial Attention Module with input vectors, rather than the sequential connection of CAM and SAM. Bearing faults often manifest in specific regions of the signal, such as particular time intervals or frequency bins. This framework can effectively obtain a large amount of local detail information from high-resolution input feature maps, and for rolling bearing fault diagnosis, these local information contain important information about the specific location and type of defect. Spatial attention allows the model to focus on these critical regions, enhancing the precision of fault localization. Furthermore, Different types of bearing faults may have different spatial patterns. Spatial attention can adapt to these variations, allowing the model to generalize well across different fault types and operating conditions. Similar to the mechanism of CAM, spatial attention also generates relevant weights for the attention map, which is replicated in concatenation. As shown in Figure 1, the parallel arrangement of channels and spatial modules are affected by tunable slopes (a) and (b) during the concatenation process. Compared to ordinary object detection tasks, the input information of rolling bearing fault diagnosis is more sensitive, and a small disturbance may have a significant impact on the type identification of the defect. Therefore, AHAB adopts trainable parameters to adjust the impact of attention modules on the model, avoiding environmental disturbances and small signal interference.

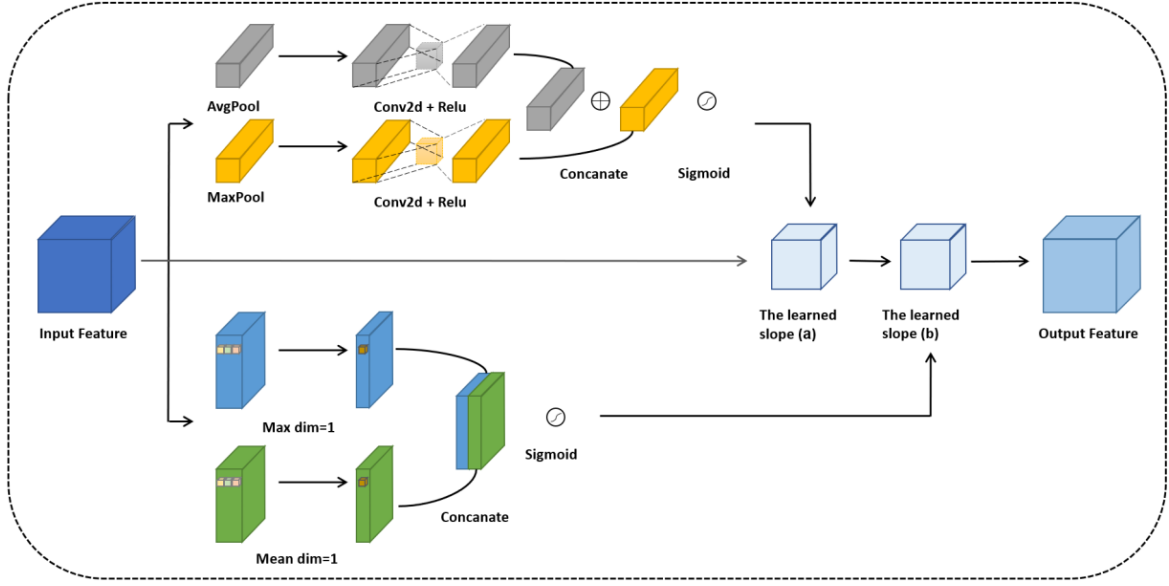

Figure 1. The diagram of AHAB architecture

The AHAB channel attention and spatial attention can be computed as follows:

$$Channel\ Attention(\mathbf{F}) = \delta\left(MLP(AvgPool(\mathbf{F})) + MLP(MaxPool(\mathbf{F}))\right) \quad (1)$$

$$\mathbf{F_C} = \alpha * Channel\ Attention(\mathbf{F}) \otimes \mathbf{F} \quad (2)$$

$$Spatial\ Attention(\mathbf{F}) = \delta(f^{7*7}([AvgPool(F); MaxPool(\mathbf{F})])) \quad (3)$$

$$\mathbf{F_S} = \beta * Spetial\ Attention(\mathbf{F}) \otimes \mathbf{F'} \quad (4)$$

The output of submodules is calculated as the combination of Multilayer Perceptron (MLP), pooling layer and activation function $\delta$, and a tunable parameter is added for each attention mechanism named as $\alpha$ and $\beta$ to realize adaptive learning. $F$ denotes the input vector. $\mathbf{F_C}$ and $\mathbf{F_S}$ represent the outputs of channel attention module and Spatial attention module respectively. $f^{7\times 7}$ in spatial attention represents a $7 \times 7$ convolution operation.

## 2.2 Residual Multi-layer Neural Network module

Feed-Forward-Network (FFN) is integrated into Single-Head Vision Transformer (SHViT) architecture. The Depthwise Convolution (DWConv) layer, Single-Head Self-Attention (SHSA) layer and the Feed-Forward-Network (FFN) are the three main components of the architecture of the original SHViT. The structure diagram of FFN is illustrated in Figure 2.

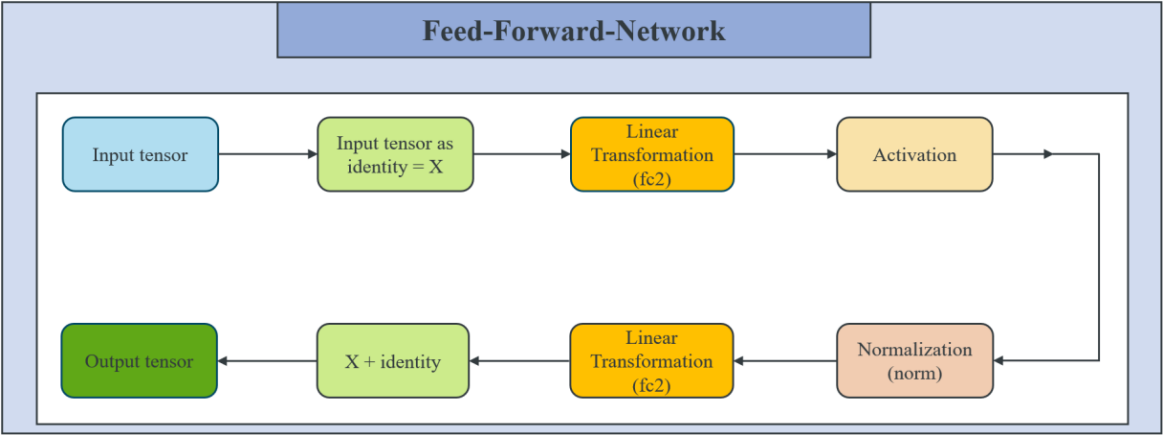

Figure 2: The structure diagram of Feed-Forward-Network (FFN)

The relationships among different patches in Vision Transformers are captured by the self-attention mechanism, while the information for each patch is processed by the FFN individually. In the Single-Head Vision Transformer (SHViT) architecture, as shown in Figure 3, we replace the traditional Feed-Forward-Network (FFN) with a Residual Feed-Forward-Network (Res-FFN) within each SHViT block. Inspired by the success of ResNet [23], the Res-FFN module enhances the model's ability to learn complex representations and mitigates the vanishing gradient problem by adding residual connections between the input and output of Res-FFN module. This modification is a key feature of our proposed framework, as illustrated in Figure 3.

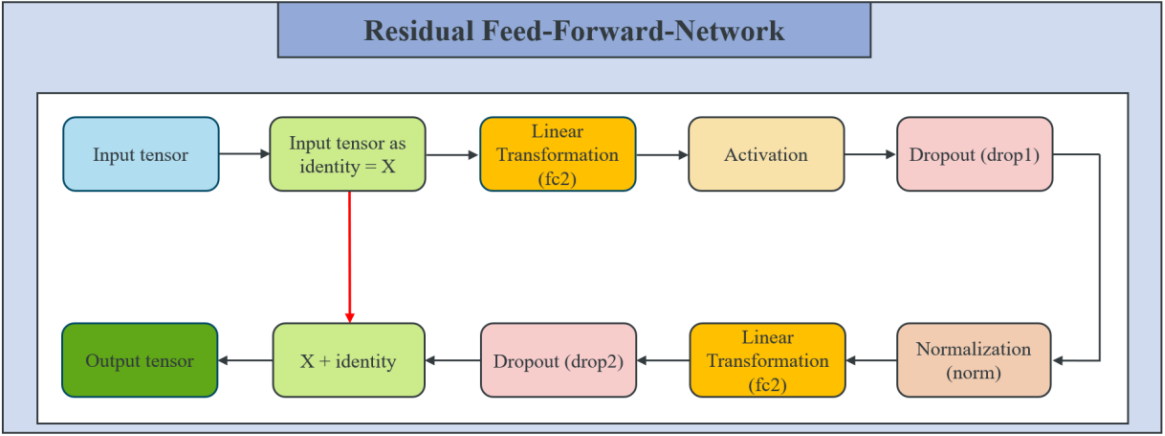

Figure 3: The structure diagram of Residual Feed-Forward-Network (Res-FFN)

Residual connections within the Res-FFN module are employed by the Residual Attention Single-Head Vision Transformer Network(RA-SHViT-Net) to address the vanishing gradient problem, enabling efficient and stable training for deep architectures. These connections allow the model to skip over less informative layers, enhancing its ability to learn more complex representations. The residual feedforward network (Res-FFN) is also configured with an expansion ratio to scale up the vector dimensions by the specified multiple. Furthermore, the designed residual connections provide robustness against variations in model hyperparameters and architectural configurations, thereby simplifying the development process of our model and accelerating model training. These enhancements contribute to faster training convergence and higher accuracy on rolling bearing data for fault diagnosis. Therefore, the Res-FFN module improves the expressiveness and generalization of the proposed model, making it more adept at capturing non-linear feature relationships. This highlights the effectiveness of the Res-FFN module in both training and generalization. Furthermore, the dropout operation is added in each layer to enhance the generalization of the network by preventing overfitting, encouraging diverse feature learning, and improving robustness to noise, ultimately helping the model perform better.

## 2.3 Residual Attention Single-Head Vision Transformer Network

To address the computational redundancies and inefficiencies associated with multi-head self-attention (MHSA) mechanisms in vision transformers, we utilize the Single-Head Vision Transformer (SHViT) as the backbone of the proposed network. SHViT is designed to streamline both training and inference processes while maintaining high performance. The entire structure diagram of RA-SHViT-Net is illustrated in Figure 4. Initially, the time-domain signals collected from sensors or from a dataset are transformed into frequency-domain using FFT. Then, the real and imaginary parts of the frequency-domain signal are concatenated along the channel direction. The data points are listed in rows and columns to form 2D-matrices, similar to an image, with the height and width being (64, 32), enhancing the distinction between the different fault types of bearings by amplifying their differences. This transformation facilitates clearer separation of fault types by revealing distinct patterns in both the frequency and time domains. These generated 2D-matrices are subsequently input into our proposed model, the Residual Attention Single-Head Vision Transformer Network (RA-SHViT-Net). As illustrated in Figure 4, the architecture of RA-SHViT-Net consists of three primary modules: Overlapping patch embedding, deep feature extraction, and fault diagnosis. The overlapping patch embedding layer is composed of four 3×3 strided convolution layers for local representations extraction. The deep feature extraction module employs three stages of stacked Residual Attention Single-Head Vision Transformer (RA-SHViT) Blocks to capture hierarchical representation, additional channel features, and spatial attention-aware features, complemented by an Adaptive Hybrid Attention Block (AHAB). Furthermore, the fault diagnosis process is composed of Multilayer Perceptrons (MLPs) for the final fault category classification of rolling bearings.

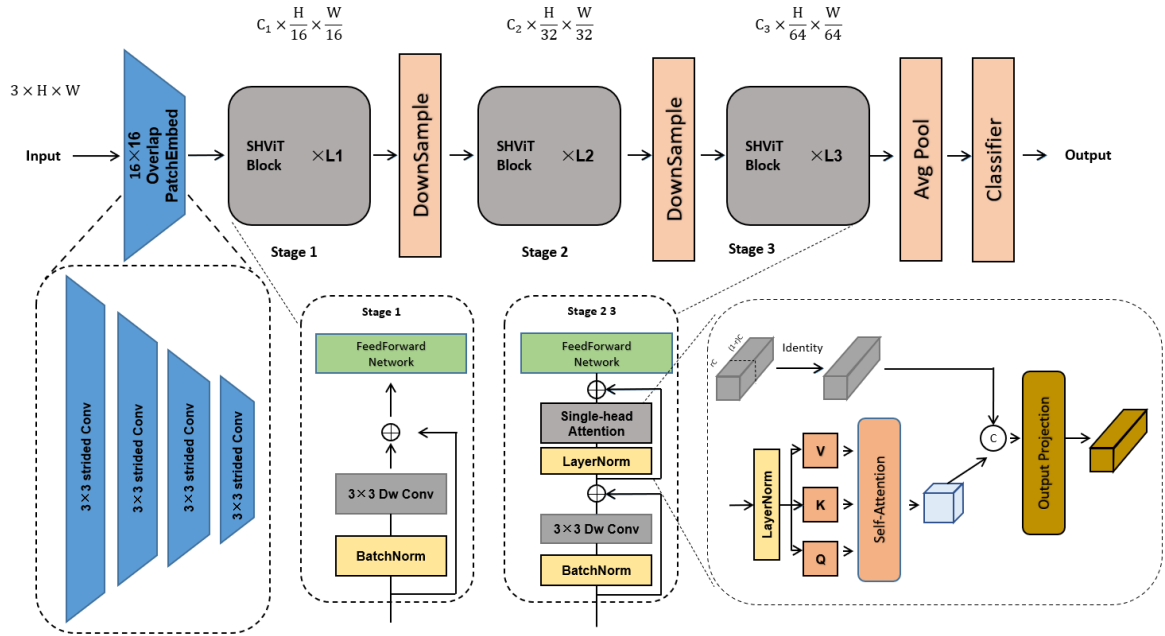

Figure 4: Residual Attention Single-Head Vision Transformer Network (RA-SHViT-Net)

    The Residual Attention Single-Head Vision Transformer (RA-SHViT) Block consists of a Depthwise Convolution (DWConv) layer, a Single-Head Self-Attention (SHSA) layer, a Residual Feed-Forward Network (Res-FFN) and an Adaptive Hybrid Attention Block (AHAB). It features a skip connection, along with a long skip connection that aggregates shallow and deep features before passing them to the fault diagnosis module, creating a "residual-in-residual" framework. Additionally, following the original SHViT model, we did not include SHSA in the RA-SHViT block during Stage 1, while SHSA was applied in Stages 2 and 3. The embedding dimension for the different stages of Residual Attention Single-Head Vision Transformer (RA-SHViT) Block is sequentially set at (128, 224, 320), with each SHSA having default partial ratio $r$ of $1/4.67$. Furthermore, the expansion ratios in the Residual Feed-Forward-Network (Res-FFN) are set to 2. The structure of STL is illustrated in Figure 5.

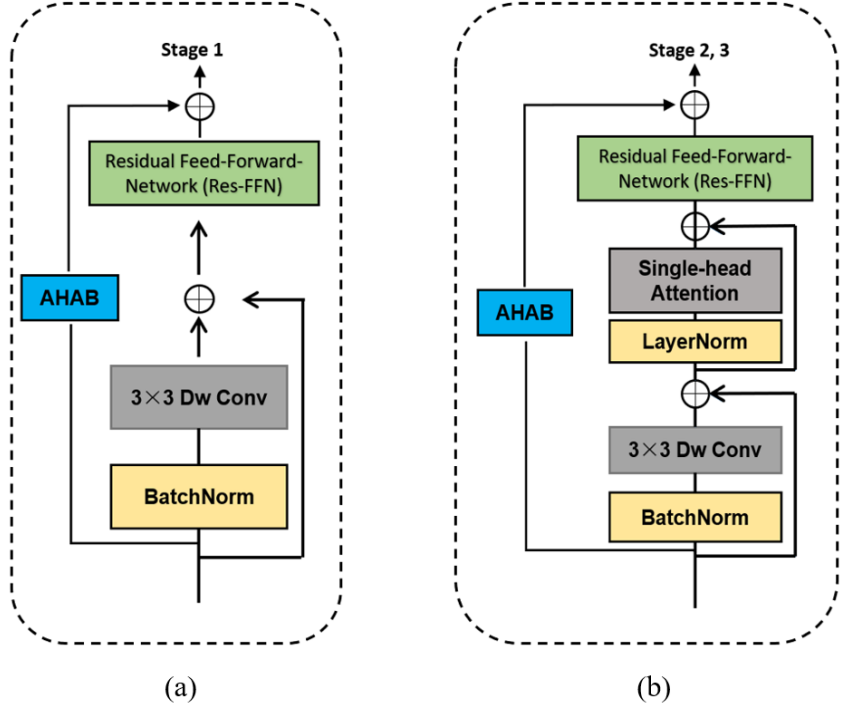

Figure 5: Structure of the Single-Head Vision Transformer (SHViT) Block in (a) Stage 1. (b) Stage 2 and 3.

The core innovation of SHViT lies in its Single-Head Self-Attention (SHSA) module, which significantly reduces computational overhead and memory usage without compromising the model's ability to capture global dependencies. In the Single-Head Vision Transformer (SHViT), the input 2D matrices are initially processed by an overlapping patch embedding stem, which comprises four 3×3 strided convolution layers. The resulting tokens are then passed through three stages of stacked SHViT blocks to extract hierarchical representations. Each SHViT block consists of three main modules (as illustrated in Figure 5): a Depthwise Convolution (DWConv) layer, a Single-Head Self-Attention (SHSA) layer, and a Residual Feed-Forward Network (Res-FFN). The DWConv layer reduces computational complexity and extracts channel-wise features, contributing to local feature aggregation or conditional position embedding. Following the DWConv layer, the output tensor is processed through the SHSA module, which models global contexts. Finally, the Res-FFN module enhances channel interactions, further refining the feature representations. The integration of DWConv layer and SHSA module effectively captures both local and global dependencies in a computationally and memory-efficient manner. The DWConv layer contributes to local feature aggregation and conditional position embedding. Subsequently, the output tensor $X$ is processed through the Single-Head Self-Attention (SHSA) module.

In SHSA module, the tensor $X$ is split into two parts: $X_{att}$, a subset of the input channels used for attention computation, and $X_{res}$, the remaining channels that are left unchanged. The number of channels in $X_{att}$ is $C_p = rC$, where $r$ is a predefined ratio (default set to 1/4.67). The subset of channels $X_{att}$ is projected into query ($Q$), key ($K$), and value ($V$) matrices using learnable weight matrices $W_Q$, $W_K$, and $W_V$:

$$X_{att}, X_{res} = Split(X, [C_p, C - C_p]) \quad (5)$$

$$Q = X_{att}W_Q, K = X_{att}W_K, V = X_{att}W_V \quad (6)$$

The attention scores are computed using the dot product between the query and key matrices, followed by a softmax normalization and scaling by $\sqrt{d_{qk}}$ (where $d_{qk}$ is the dimension of the query and key, typically set to 16). The result of the attention mechanism is the attended features $\tilde{X}_{att}$. Next, these attended features are concatenated with the residual features $X_{res}$ along the channel dimension. Finally, the concatenated tensor is projected back to the original channel dimension using a learnable weight matrix $W_O$:

$$Attention(Q, K, V) = Softmax\left(\frac{QK^T}{\sqrt{d_{qk}}}\right)V \quad (7)$$

$$\tilde{X}_{att} = Attention(Q, K, V) \quad (8)$$

$$SHSA(X) = Concat(\tilde{X}_{att}, X_{res})W_O \quad (9)$$

The output of the SHSA module is then passed through a Residual Feed-Forward-Network (Res-FFN) to further refine the feature representations. This process is iterated across multiple layers, enabling SHViT to construct a deep hierarchical understanding of the input data. Each layer refines the feature representations, capturing both local and global dependencies. The expansion ratios in the Residual Feed-Forward-Network (Res-FFN) are set to 2, enhancing the model's capacity to learn complex features. The integration of Depthwise Convolution (DWConv) and Single-Head Self-Attention (SHSA) efficiently captures both local and global dependencies while minimizing computational and memory requirements. Notably, the SHSA layer is omitted in the initial stage to optimize resource utilization. To reduce the number of tokens without information loss, an efficient downsampling layer is employed, consisting of two stage 1 blocks separated by an inverted residual block with a stride of 2. Finally, the global average pooling and fully connected layer are utilized to generate predictions.

## 3 EXPERIMENTS

### 3.1 Data description

This section describes the experimental platform and data set setup and provides a comparative analysis of the experimental results. The proposed model designed in the third section is programmed in Python based on the PyTorch 1.13 back-end language, running on Ubuntu 20.04 and Nvidia A100 40 GB x 2 GPU.

#### 3.1.1 The Case Western Reserve University (CWRU) Bearing Dataset

The Case Western Reserve University (CWRU) dataset uses vibration signals from an SKF 6205-2RS rolling bearing, which sampled at 12 kHz, includes four operational conditions and covers single-point faults on the ball, inner race, and outer race, with fault diameters of 0.18 mm, 0.355 mm, and 0.533 mm, respectively. Additionally, data from normal bearings were included. Table 1 provides a detailed overview of the ten distinct fault types examined in the study.

Table 1: Overview of the CWRU bearing dataset

| Label | Fault Type | Fault Diameter |
|---|---|---|
| 1 | Rolling Element Fault | 0.18 |

| Label | Fault Type | Fault Diameter |
|---|---|---|
| 2 | Rolling Element Fault | 0.355 |
| 3 | Rolling Element Fault | 0.533 |
| 4 | Inner Ring Failure | 0.18 |
| 5 | Inner Ring Failure | 0.355 |
| 6 | Inner Ring Failure | 0.533 |
| 7 | Outer Ring Failure | 0.18 |
| 8 | Outer Ring Failure | 0.355 |
| 9 | Outer Ring Failure | 0.533 |
| 10 | Normal | 0 |

#### 3.1.2 The Paderborn University (PU) Bearing Dataset

The Paderborn bearing dataset is also utilized in this study. Its modular test bench consists of five main components: a motor, a torque measurement shaft, a rolling bearing test module, a flywheel, and a load motor. The experiment centers on 6203-type rolling bearings. Vibration data for the bearings were sampled at 64 kHz over intervals of 4 seconds each. Specifically, the N15_M07_F04 dataset was chosen for validation, containing both baseline signals and human-induced fault conditions. During testing, the system operated at a speed of 1500 revolutions per minute, with a torque of 0.7 Newton meters and a radial force of 400 Newtons. Artificial faults were created on both the inner and outer rings, employing various machining techniques. Table 2 provides a detailed overview of the failure sample set.

Table 2: Overview of the PU bearing dataset

| NO. | Bearing Code | Fault Mode | Description |
|---|---|---|---|
| 0 | KA04 | Outer ring damage | Caused by fatigue and pitting |
| 1 | KA15 | Outer ring damage | Caused by plastic deform and indentation |
| 2 | KA16 | Outer ring damage | Caused by fatigue and pitting |
| 3 | KA22 | Outer ring damage | Caused by fatigue and pitting |
| 4 | KA30 | Outer ring damage | Caused by plastic deform and indentation |
| 5 | KB23 | Outer ring and innerring damage | Caused by fatigue and pitting |
| 6 | KB24 | Outer ring and innerring damage | Caused by fatigue and pitting |
| 7 | KB27 | Outer ring and innerring damage | Caused by plastic deform and indentation |
| 8 | KI14 | Inner ring damage | Caused by fatigue and pitting |
| 9 | KI16 | Inner ring damage | Caused by fatigue and pitting |
| 10 | KI17 | Inner ring damage | Caused by fatigue and pitting |
| 11 | KI18 | Inner ring damage | Caused by fatigue and pitting |
| 12 | KI21 | Inner ring damage | Caused by fatigue and pitting |
| 13 | KI04 | Inner ring damage | Caused by fatigue and pitting |

### 3.2 Experiment setting

In this study, data were selected from drive end bearings operating at a motor speed of 1790 rpm with a load of 0 hp, and a sampling frequency of 12 kHz. Bearing faults are classified into ten categories based on fault location and size. Each sample from the fault dataset is a fixed-length signal segment containing 2048 data points, subsequently converted into 64 by 32-pixel images, with 2000 samples collected for each bearing state. The CWRU dataset is divided into training, validation, and test sets with a 7:1:2 ratio for the experiments. For the PU dataset experiments, we selected 250 training samples and 250 test samples for each of the six states (K001, KA01, KA03, KA07, KI01, KI03), ensuring an equal 1:1 ratio between training and test samples. Data in this study were gathered through sliding window sampling. To address the risk of overfitting caused by limited training data, data augmentation strategies were applied to expand the training set significantly. The issue of limited training data is effectively mitigated in this case study using a sliding overlapping sampling approach. The primary parameters for the training processes are set as follows: a learning rate of 0.001, a batch size of 16, 750 epochs, and the AdamW optimizer. The specific hyperparameters setting for the RA-SHViT-Net model have already been detailed in Sec 2.3.

### 3.3 Performance in noisy environments

In actual working environment, the collection of bearing vibration data is interfered with by noise. To facilitate the restoration of conditions under which our model operates in different noisy environments, the raw bearing vibration signals from the CWRU dataset are initially normalized. Subsequently, Gaussian white noise at different signal-to-noise ratios (SNRs) is added to the normalized signals. The SNR is defined as follows:

$$SNR = 10 \times \log_{10}\left(\frac{P_{signal}}{P_{noise}}\right) \quad (10)$$

In this study, the signal strength of the original bearing vibration is denoted as $P_{signal}$, while $P_{noise}$ represents the strength of the added noise. The SNR is calculated as the ratio of the average power of the original vibration signal to the average power of the noise signal, providing a measure of signal quality. Gaussian white noise with specific SNR values was introduced to the original vibration signal. Gaussian white noise, with the specific magnitude, is then added to the original signal, resulting in a noisy signal. Subsequently, the time-domain signals from this dataset were transformed into frequency-domain signals using FFT. The training set was used for model training, while the performance of our proposed model under different levels of noise was assessed with the test set.

### 3.4 Comparative experiments and Analysis

#### 3.4.1 Results of CWRU Bearing Dataset.

The training and validation of multiple experiments were conducted independently, with the comparative results presented in Figure 6. Our model achieves an accuracy of 69.7% on the test set under a noise level of -10dB. Furthermore, it achieves 99.8% accuracy under a 10dB noise condition, with only 10 samples being incorrectly classified overall. Our proposed model attains the highest average accuracy of 93.5%, outperforming TAR [24], Transformer [17], TCN [25], WDCNN [26], and GRU [27] by 1.9%, 13.6%, 9.9%, 13.6%, and 22.0%, respectively, across SNRs ranging from -10dB to 10dB. As the SNR decreases, the average diagnostic accuracy of all algorithms declines, albeit to varying degrees. The proposed method in this study consistently outperforms other algorithms in all SNR conditions (especially low SNR conditions), highlighting its robustness in high-noise environments. These findings show the proposed RA-SHViT-Net is more effective in handling samples in high-noise scenarios.

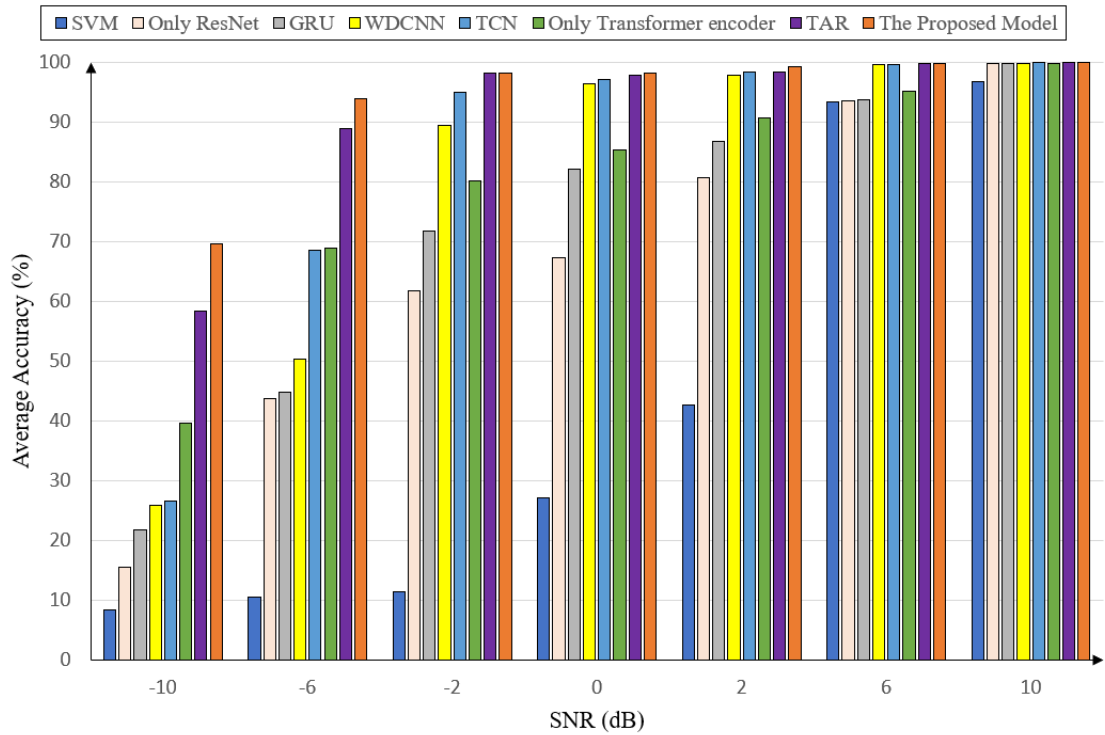

Figure 6: Comparison of prediction accuracies of different models on the CWRU bearing dataset

The prediction results were analyzed using a confusion matrix to address the potential for extreme misidentification of specific fault types. Figure 7(a) illustrates the relatively poor classification performance of the model in a -10dB noise environment, particularly where numerous samples from label 4 were misclassified as label 2. However, this extremely minor misclassification is illustrated in Figure 7(d). The method demonstrates high precision and stability across various noise environments. Additionally, in high-noise environments, the diagnostic performance for the 0.18 mm inner race fault is somewhat diminished, but the overall accuracy remains close to or exceeds 70%.

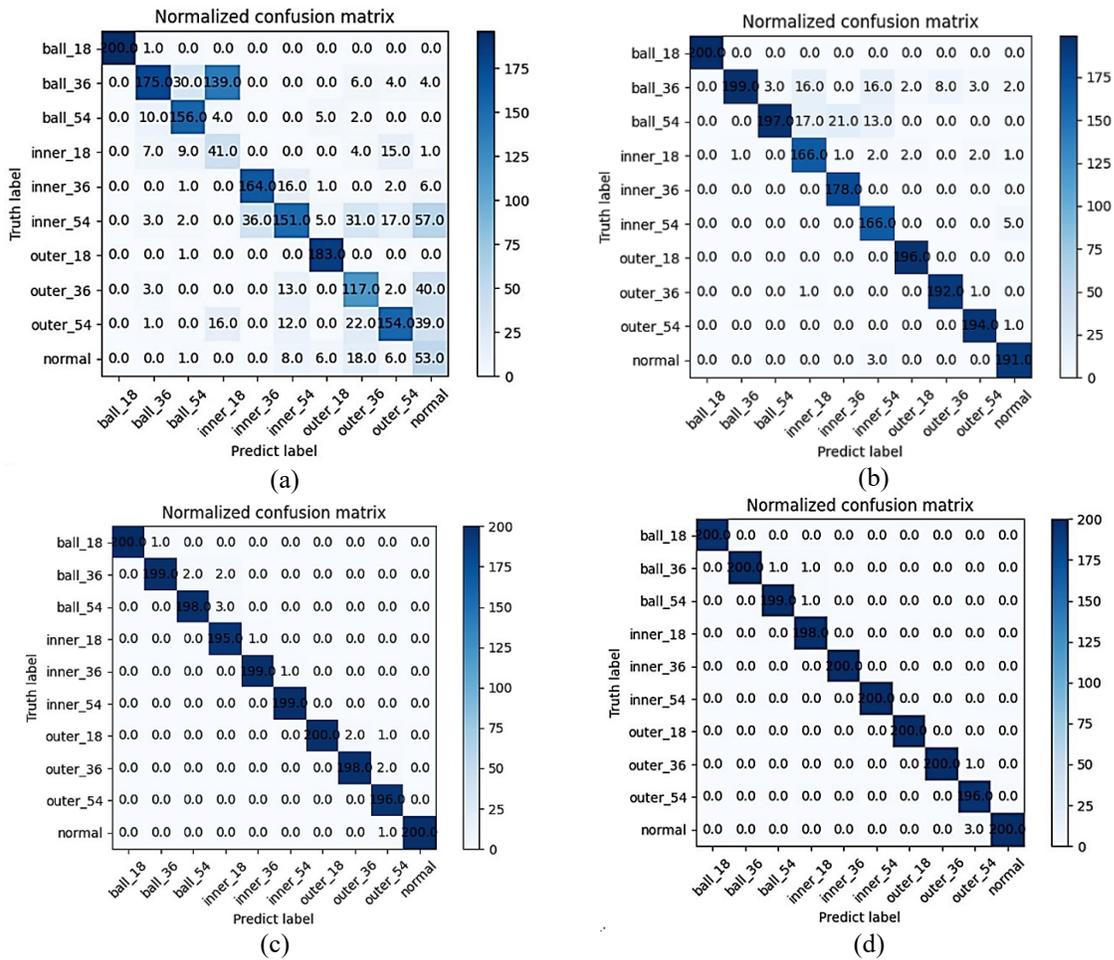

Figure 7: Classification confusion matrix for prediction results at different levels of noise on the CWRU bearing dataset. (a) -10dB. (b) -6dB. (c) 2dB. (d) 6dB

The high-dimensional features are first reduced using t-SNE [28] to facilitate visualization of the training feature vectors, and the results are demonstrated in Figure 8. Among these visualizations, more sample types overlap under -10dB noise environment, while only several sample types overlap under the 6dB noise environment. The clustering effect in the experiments is significantly enhanced as the value of SNR increasing. This improvement demonstrates the superior clustering capability of our proposed model.

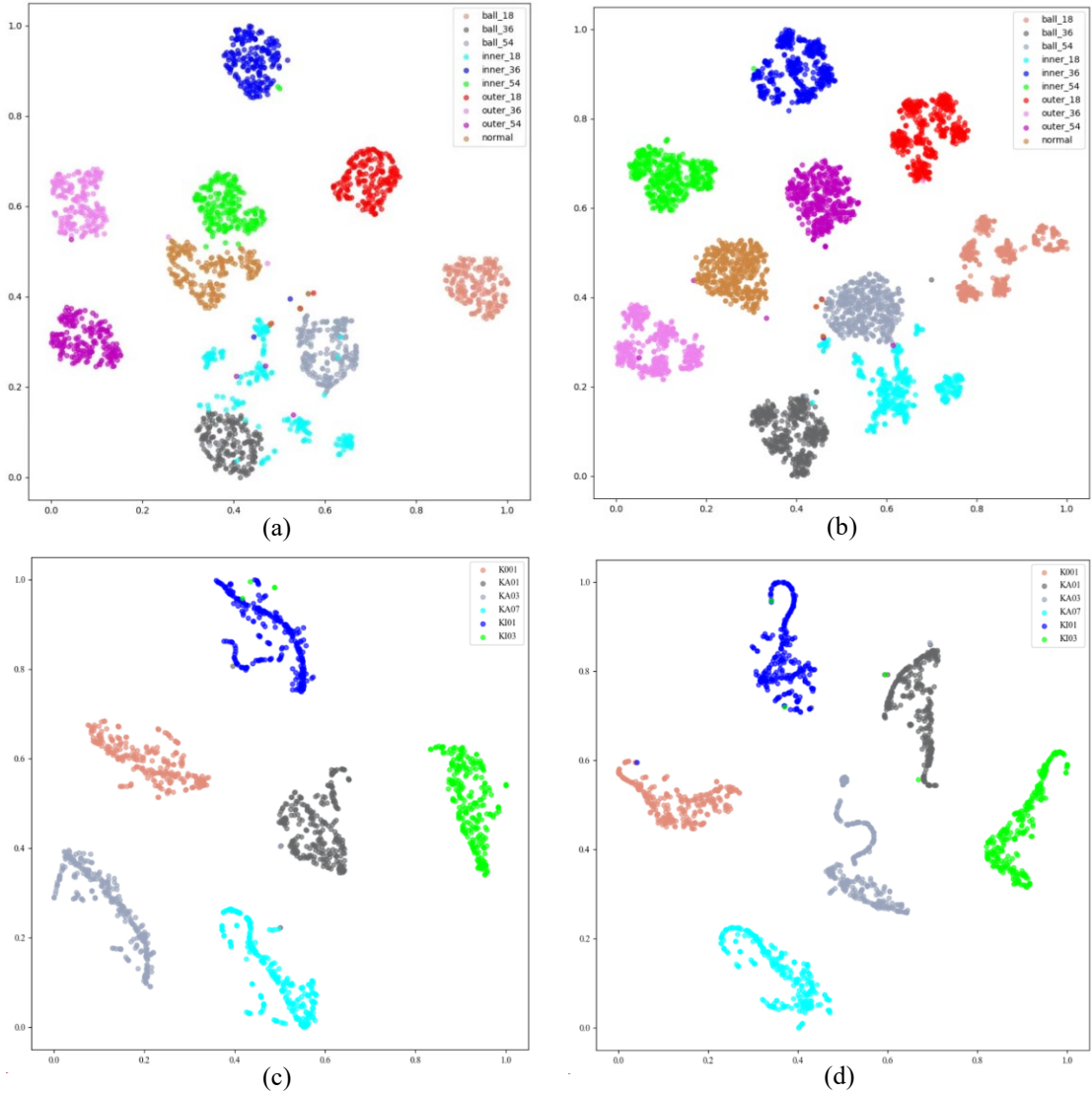

Figure 8: Clustering visualization for prediction results at different levels of noise on the CWRU bearing dataset. (a) -10dB. (b) -6dB. (c)2dB. (d) 6dB

### 3.4.2 Results of PU Bearing Dataset.

Independent training and validation were conducted for multiple experiments, with the comparative results shown in Figure 9. Our model achieves 96.9% accuracy on the test set at a noise level of -4dB and 99.7% accuracy at 4dB noise, with only 4 samples misclassified overall. The proposed model attains the highest average accuracy, 99.1%, surpassing MARP [29], VSI-DGGAPN [30], HMCNN [31], and SVM [32] by margins of 0.42%, 23.54%, 6.79%, and 25.19%, respectively, across SNRs from -4dB to 4dB. As SNR decreases, the average diagnostic accuracy of all algorithms declines to varying extents. Nevertheless, the proposed method consistently outperforms other algorithms across all SNR conditions, particularly in low-SNR environments, underscoring its robustness in high-noise settings. These results demonstrate that the proposed RA-SHViT-Net is especially effective for sample classification in high-noise scenarios.

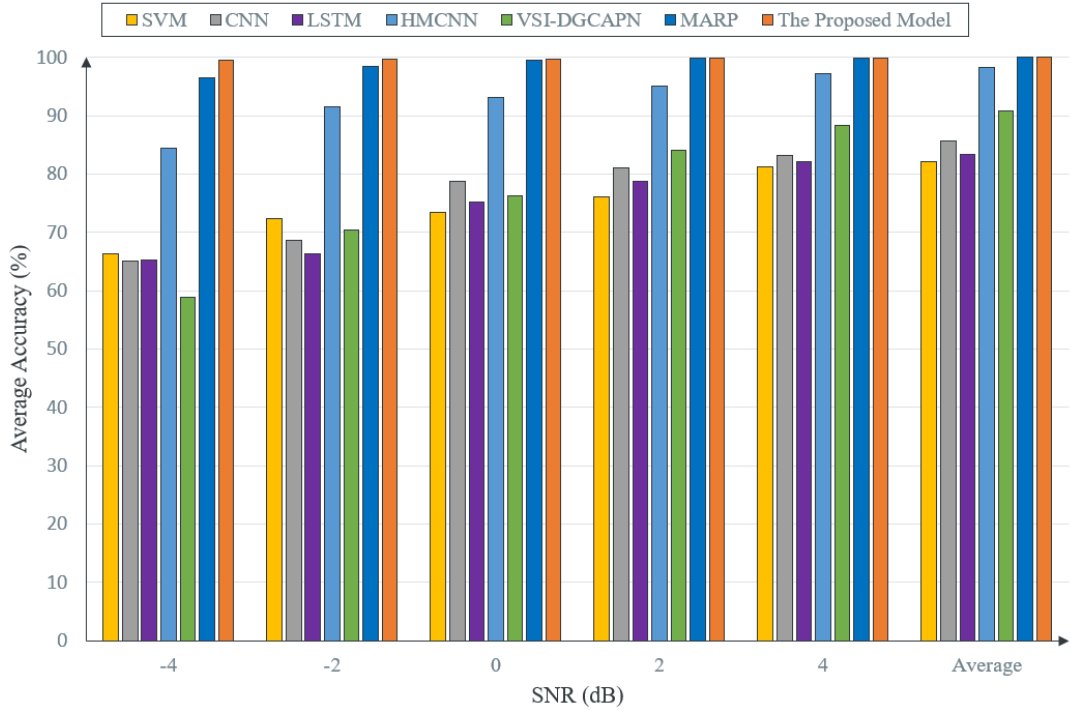

Figure 9: Comparison of prediction accuracies of different models on the PU bearing dataset

A confusion matrix was employed to analyze the prediction results, focusing on the potential for significant misclassification of certain fault types. As shown in Figure 10(a), the model's classification performance is relatively lower in a -4dB noise environment. In contrast, Figure 10(d) reveals only minimal misclassification under a 4dB noise environment. Overall, the method demonstrates strong precision and stability across different noise levels.

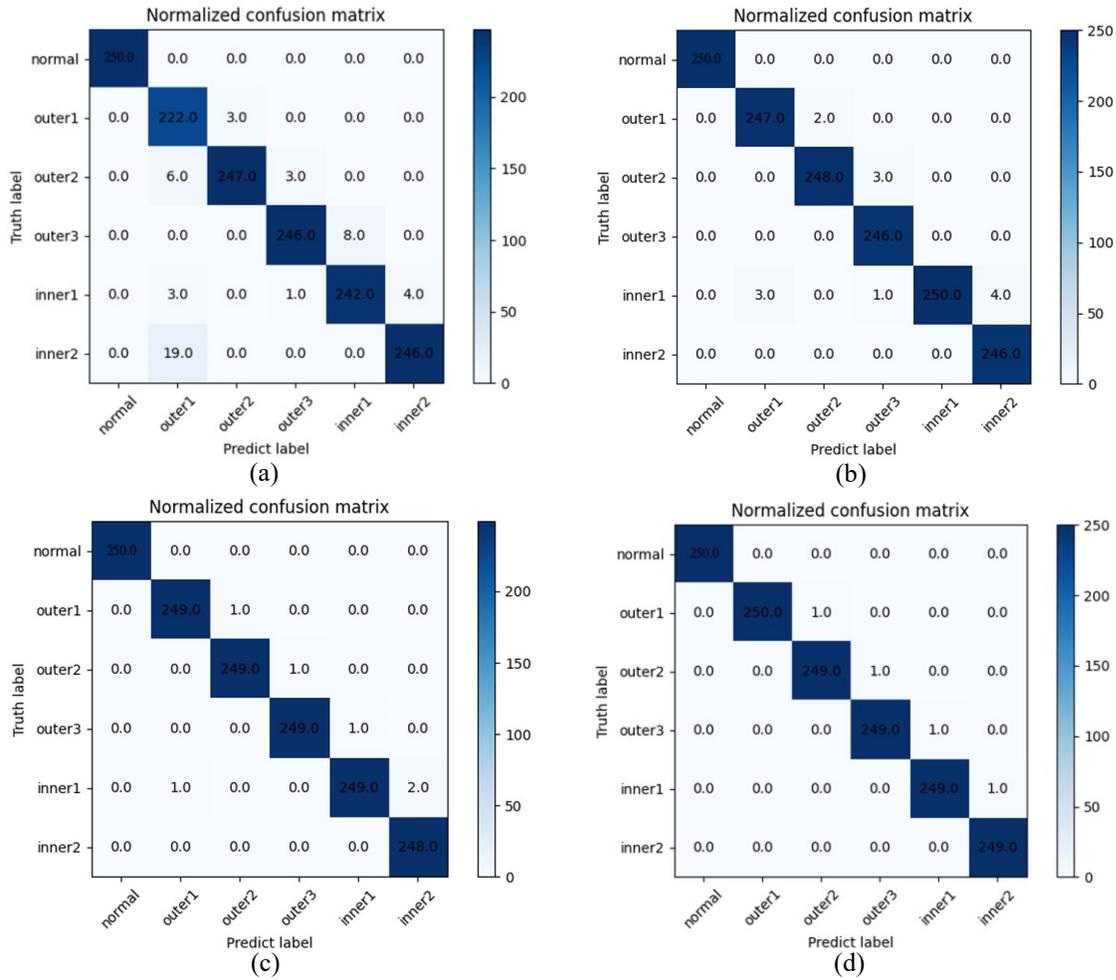

Figure 10: Classification confusion matrix for prediction results at different levels of noise on the PU bearing dataset. (a) -4dB. (b) -2dB. (c) 2dB. (d) 4dB

Similarly, t-SNE [28] is applied to visualize the training feature vectors, with results displayed in Figure 11. Under a -4dB noise environment, more sample types exhibit overlap, whereas under a 4dB noise environment, only a limited number of sample types overlap. The clustering effect improves significantly as the SNR value increases, highlighting the strong clustering capability of the proposed model.

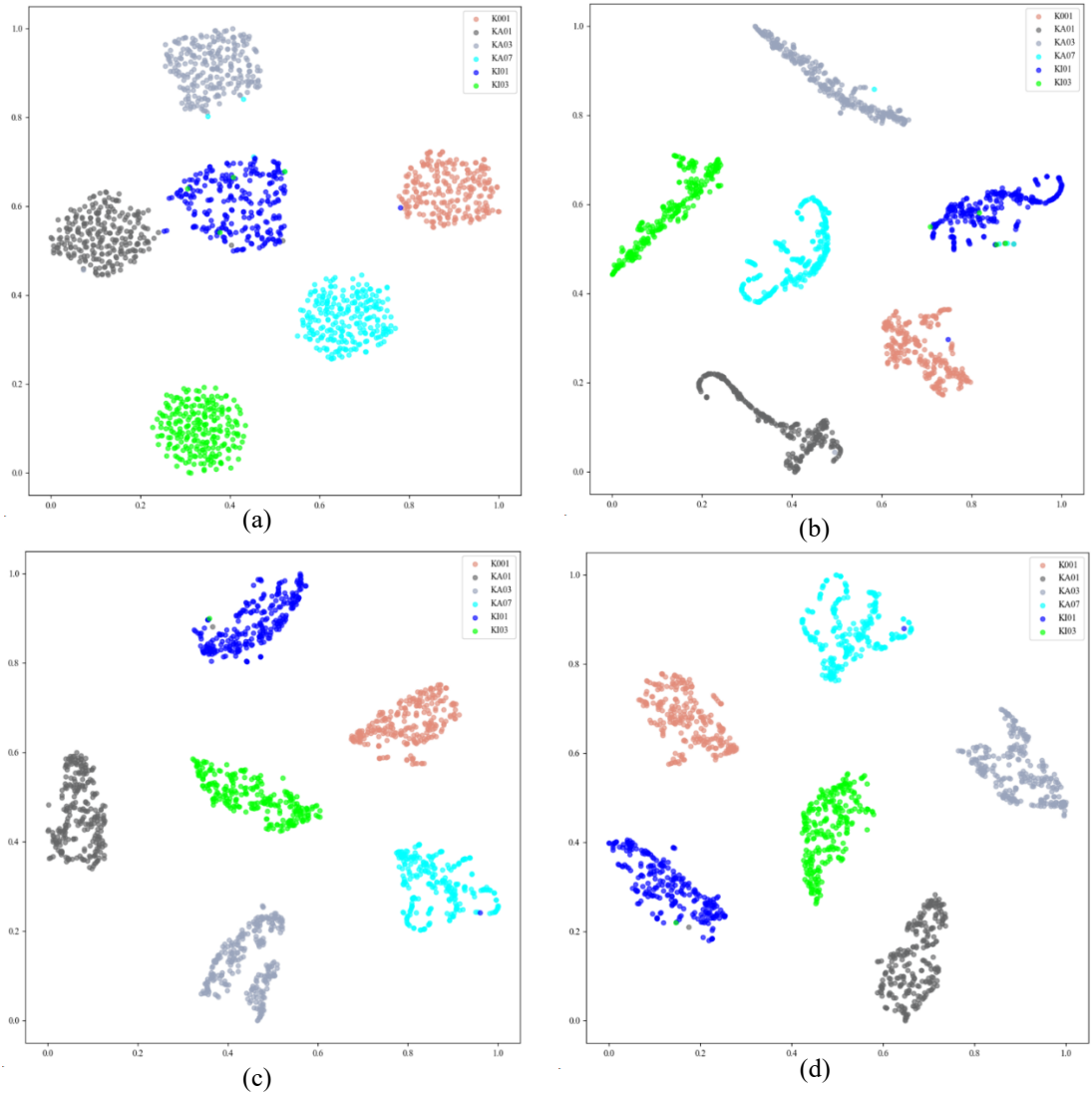

Figure 11: Clustering visualization for prediction results at different levels of noise on the PU bearing dataset. (a) -4dB. (b) -2dB. (c) 2dB. (d) 4dB

### 3.4.3 Results of CWRU Bearing Dataset.

Additionally, we designed experiments to investigate the lightweight structure of the SHViT model to assess whether RA-SHViT-Net achieves a state-of-the-art balance between computational complexity and predictive accuracy compared with other previous Transformer networks. For this purpose, we selected the most representative models, ViT [33] and Swin-T [34], for comparison and the experiments are conducted in the 0dB noise environment. All other experimental settings remain consistent with those previously described. The final experimental results are presented in Table 3.

Table 3: Performance comparisons of various Transformer series networks

| Network | Accuracy (%) | Flops (M) | Params (M) |
|---|---|---|---|
| ViT | 96.3 | 33.52 | 42.37 |
| Swin-T | 97.6 | 27.49 | 35.70 |
| RA-SHViT-Net | 98.2 | 6.01 | 19.46 |

Therefore, from Table 3, we can see that RA-SHViT-Net not only outperforms ViT and Swin-T in prediction accuracy, but also uses fewer flops and parameters. Therefore, we can conclude that SHViT does provide the most advanced balance between computational complexity and prediction accuracy.

### 3.5 Ablation Study

#### 3.5.1 Study the impact of AHAB.

This section mainly studies the impact of AHAB on our proposed entire network. Multiple sets of experiments are designed for comparison under different noise levels. The models used in all experiments retain the same structure-control variables except whether AHAB is configured. All other experimental settings remain consistent with those previously described. All experiments are conducted using the CWRU dataset. The results are shown in Figure 12.

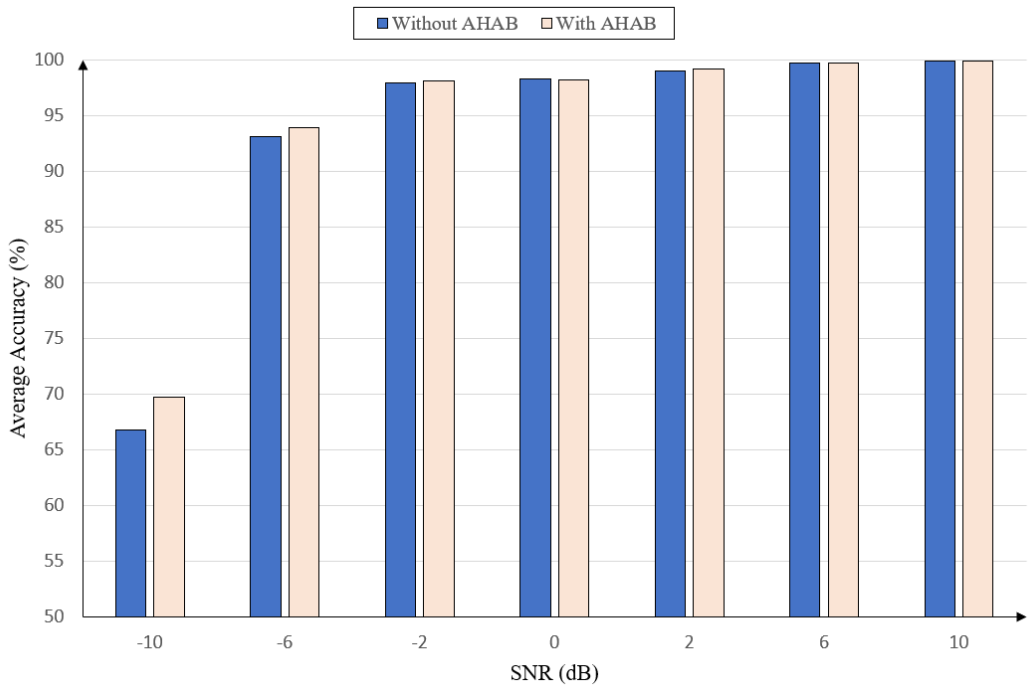

Figure 12: Prediction results of our proposed network with or without AHAB.

The results suggest that networks incorporating AHAB demonstrate significantly improved predictive performance in high-noise environments compared to those without AHAB. For instance, in a noise environment of -10 dB, the prediction accuracy of the network without AHAB is 66.8%, whereas the network with AHAB achieves an accuracy of 69.7%. As noise decreases, the performance gap between the two networks progressively narrows, with the difference becoming nearly negligible in very low-noise settings. This confirms AHAB's substantial benefit in enhancing model prediction accuracy under high-noise conditions, while its impact is minimal in low-noise environments.

### 3.5.2 **Study the impact of Res-FNN.**

This section mainly studies the impact of Res-FNN on our proposed entire network. Multiple sets of experiments are also designed for comparison under different noise levels. The models used in all experiments retain the same structure-control variables except whether Res-FNN or original FNN is configured. All other experimental settings remain consistent with those previously described. All experiments are conducted using the CWRU dataset. The results are shown in Figure 13.

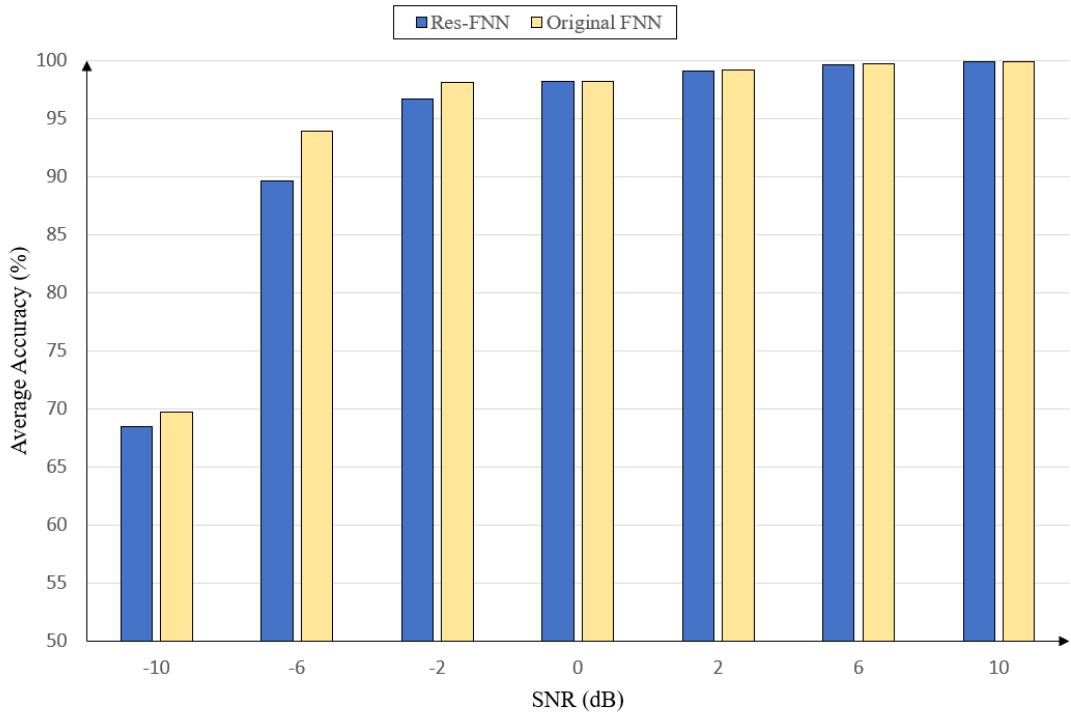

Figure 13: Prediction results of our proposed network with Res-FNN or original FNN.

It can be summarized from the result diagram that the prediction accuracy of the network configured with Res-FNN in a -10dB noise environment is 69.7, while the prediction accuracy of the network configured with original FNN in a -10dB noise environment is 68.5, which is worse than that of the network configured with Res-FNN. As the external noise weakens, the prediction accuracy of the network configured with Res-FNN improves much faster than the network configured with original FNN. In a high-noise environment of -6dB, the prediction accuracy of the network configured with Res-FNN is 93.9, which is significantly higher than the prediction accuracy of 89.6 of the network configured with original FNN. Then the SNR value becomes larger, and the accuracy difference between the two becomes smaller, especially for very low noise environments, the difference basically disappears. This also proves that Res-FNN improves the model's robustness to noise and has higher expressiveness and generalization capabilities in improving the model's prediction effect.

### 3.5.3 Study the impact of FFT.

Since the previous experiments were based on the default data after FFT operation and then input to the model, in this section, we will study the impact of FFT operation on the prediction effect of the model. We divide the model into two categories: using FFT processed data and using original data for prediction and conduct multiple groups of experiments. All other experimental settings remain consistent with those previously described. All experiments are conducted using the CWRU dataset. The results are shown in Figure 14.

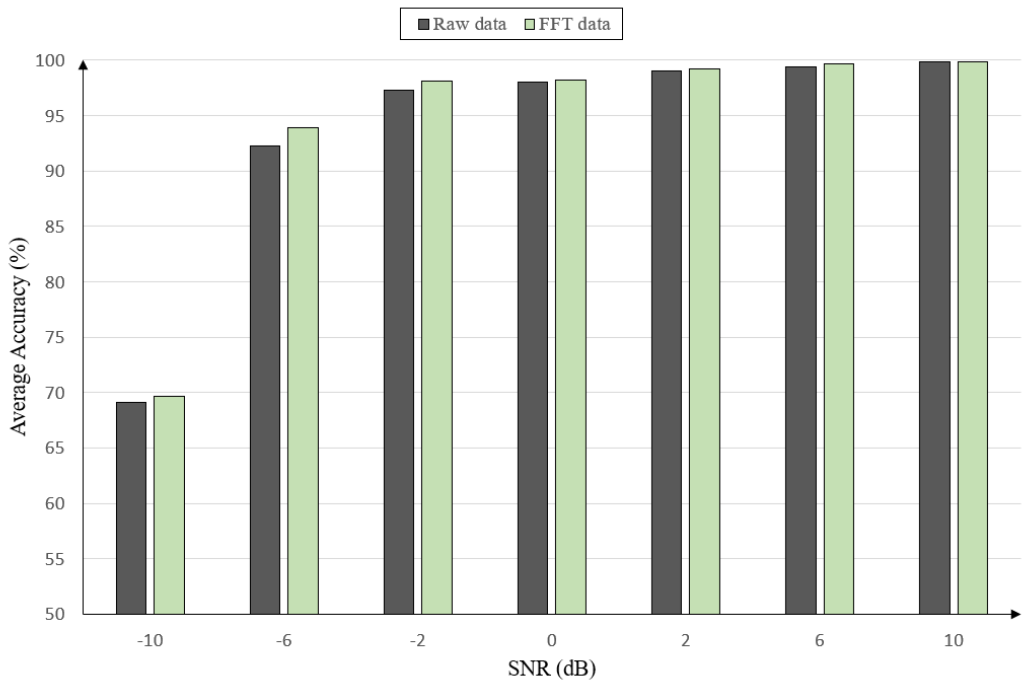

Figure 14: Prediction results of our proposed network using FFT data or Raw data.

From the above figure, we can see that in the noise environment of the entire range of -10dB to 10dB, the prediction effect of the model using FFT data is always slightly better than that of the model using raw data. This experimental result proves the effectiveness of FFT, which reveals frequency domain insights, makes characteristic fault frequencies easier to identify, and enhances the model's ability to associate specific frequencies with fault types (such as faults in the inner or outer ring). In addition, FFT helps to extract features more efficiently and generally provides a compact data representation, which can improve prediction accuracy and reduce computational complexity.

## 4 CONCLUSION

Rolling bearings are essential components in modern industrial machinery, playing a pivotal role in ensuring the performance, longevity, and safety of equipment. However, due to the harsh operating conditions such as high speeds and temperatures, rolling bearings are susceptible to malfunctions, which can lead to significant economic losses and safety risks. To address these challenges, this paper proposes the Residual Attention Single-Head Vision Transformer Network (RA-SHViT-Net) for fault diagnosis in rolling bearings, particularly in noisy environments. The RA-SHViT-Net model leverages the Single-Head Vision Transformer (SHViT) to capture both local and global features from time-series signals, offering a state-of-the-art balance between computational complexity and prediction accuracy. To enhance feature extraction, an Adaptive Hybrid Attention Block (AHAB) is introduced, combining channel and spatial attention mechanisms. This architecture is designed to comprehensively extract vibration signal features by considering the interdependence among feature channels and spatial information, thereby improving the model's robustness and accuracy. The core building block of the RA-SHViT-Net is the Residual Attention Single-Head Vision Transformer Block, which includes a Depthwise Convolution (DWConv) layer, a Single-Head Self-Attention (SHSA) layer, a Residual Feed-Forward Network (Res-FFN), and an AHAB. The Res-FFN module, with its residual connections, mitigates the vanishing gradient problem, enabling stable and efficient training. The AHAB further enhances the model's ability to focus on critical regions of the signal, improving the precision of fault localization and the model's generalization across different fault types and operating conditions. Finally, extensive experiments were conducted on the CWRU dataset and the PU dataset to validate the superior performance of the proposed model. Furthermore, multiple ablation studies were designed to examine the influence of different modules on the model's predictive performance. RA-SHViT-Net model provides a powerful tool for rolling bearing fault diagnosis, especially in noisy environments. Its ability to efficiently and accurately diagnose faults can help prevent equipment downtime, reduce economic losses, and enhance safety in industrial settings. Future work will focus on further optimizing the model and exploring its application in other types of industrial machinery.